\documentclass{article}

    \PassOptionsToPackage{numbers, compress}{natbib}

    \usepackage[preprint]{neurips_2025}

\usepackage[utf8]{inputenc} 
\usepackage[T1]{fontenc}    
\usepackage{hyperref}       
\usepackage{xurl}           
\usepackage{booktabs}       
\usepackage{amsfonts}       
\usepackage{nicefrac}       
\usepackage{microtype}      
\usepackage{xcolor}         

\usepackage{amsmath}
\usepackage{amssymb}
\usepackage{mathtools}
\usepackage{amsthm}
\usepackage{caption}
\usepackage{subcaption}
\usepackage[export]{adjustbox}%
\usepackage{multirow}
\usepackage[most]{tcolorbox}    
\usepackage{tikz}
\usetikzlibrary{shapes.geometric, positioning, arrows.meta}
\usepackage{enumitem}
\setlist[itemize]{leftmargin=*,nosep}
\usepackage{array}
\usepackage{fancyhdr}

\title{From Firewalls to Frontiers: \\
AI Red-Teaming is a Domain-Specific Evolution of Cyber Red-Teaming}

\author{%
  Anusha Sinha\thanks{Equal contribution. Correspondence to \texttt{asinha@sei.cmu.edu}} \\
  Software Engineering Institute \\
  Carnegie Mellon University \\
  \And
  Keltin Grimes\footnotemark[1] \\
  Software Engineering Institute \\
  Carnegie Mellon University \\
  \AND
  James Lucassen \\
  Independent \\
  \And
  Michael Feffer \\
  Carnegie Mellon University \\
  \And
  Nathan VanHoudnos \\
  Software Engineering Institute \\
  Carnegie Mellon University \\
  \And
  Zhiwei Steven Wu \\
  Carnegie Mellon University \\
  \And 
  Hoda Heidari \\
  Carnegie Mellon University \\
}

\begin{document}

\maketitle
\thispagestyle{fancy}   

\begin{abstract}
  A red team simulates adversary attacks to help defenders find effective strategies to defend their systems in a real-world operational setting. As more enterprise systems adopt AI, red-teaming will need to evolve to address the unique vulnerabilities and risks posed by AI systems. We take the position that AI systems can be more effectively red-teamed if AI red-teaming is recognized as a domain-specific evolution of cyber red-teaming. Specifically, we argue that existing Cyber Red Teams who adopt this framing will be able to better evaluate systems with AI components by recognizing that AI poses new risks, has new failure modes to exploit, and often contains unpatchable bugs that re-prioritize disclosure and mitigation strategies. Similarly, adopting a cybersecurity framing will allow existing AI Red Teams to leverage a well-tested structure to emulate realistic adversaries, promote mutual accountability with formal rules of engagement, and provide a pattern to mature the tooling necessary for repeatable, scalable engagements. In these ways, the merging of AI and Cyber Red Teams will create a robust security ecosystem and best position the community to adapt to the rapidly changing threat landscape.
\end{abstract}

\section{Introduction}\label{sec:intro}

Red-teaming has a long history in cybersecurity, which has developed a mature ecosystem of tools, methodologies, best-practices, and practitioners \citep{sinha2025what}. Cyber red-teaming emphasizes grounding engagements in realistic threat modeling and adversary emulation to discover the highest-risk vulnerabilities \citep{ISECOM_2012}. Cyber Red Teams \citep{cnss2015glossary} are supported by a strong culture of responsible disclosure of vulnerabilities through formal processes such as Coordinated Vulnerability Disclosure (CVD) \citep{ISECOM_2012, householder2017cert}. In contrast, AI red-teaming has only recently emerged as a popular approach to address the risks posed by AI systems \citep{ganguli2022red, feffer2024red, ahmad2024openai, shah2025approach, grattafiori2024llama}. AI red-teaming has seen extensive interest from policy-makers, researchers, and model developers, and red-teaming is quickly becoming a central security approach for many organizations \citep{biden2023executive, grattafiori2024llama, shah2025approach, bullwinkel2025lessons, openai2025preparedness, anthropic2025rsp}.

\begin{figure}[t]
\centering
\includegraphics[width=0.95\textwidth]{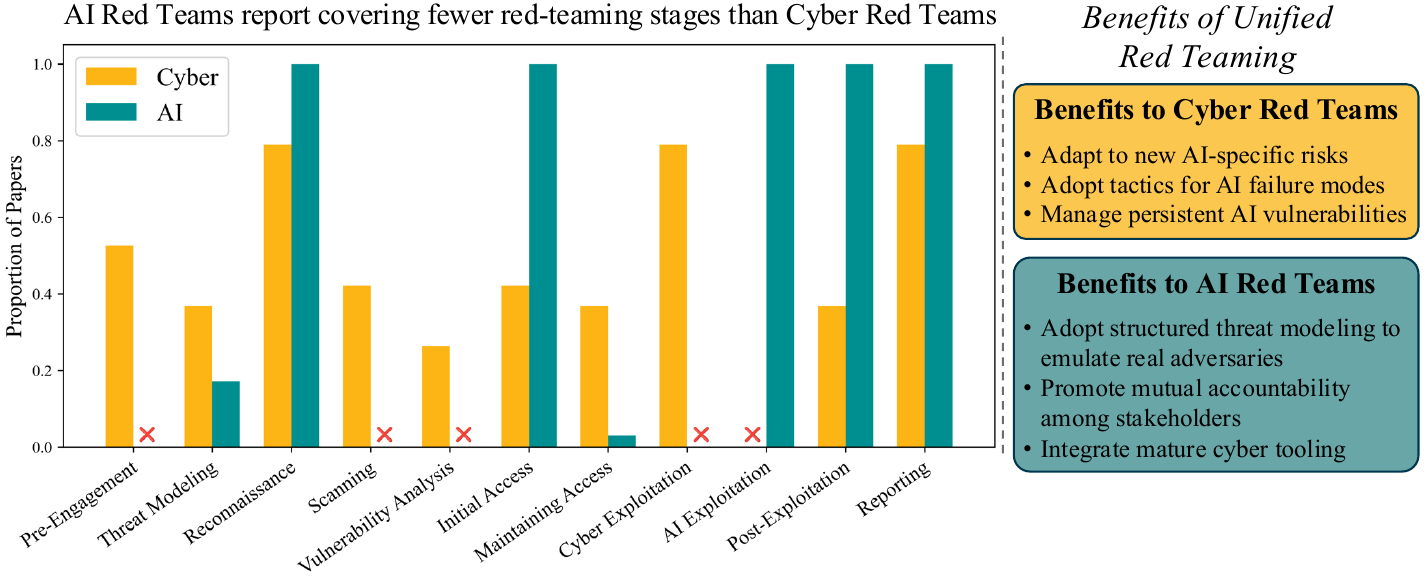}
\captionsetup{font=small}
\caption{Left: Distribution of red team stages across the AI Red Team and Cyber Red Team literature (99 and 69 papers, respectively). The gold bars display the distribution of common red team stages referenced across the Cyber Red Team papers reviewed by \citep{sinha2025what}, while the green bars display the corresponding distribution of AI Red Team papers. The red X's  display red-team stages that no papers referenced. Right: Potential benefits to Cyber Red Teams and AI Red Teams of re-framing AI red-teaming as a domain-specific evolution of cyber red-teaming.}
\label{fig:cyber_ai_comparison}
\end{figure}

Although AI Red Teams and Cyber Red Teams are both asked to address AI security concerns with the rapid growth of AI systems, a recent systematic review comparing the red-teaming literature for both AI and Cyber Red Teams found that these two communities of practice operate mostly independently, each with their own ecosystem of resources, researchers, conferences, and more \citep{sinha2025what}. This division hampers both red-teaming communities' ability to effectively respond to threats posed by AI systems. For example, Figure~\ref{fig:cyber_ai_comparison} (left) shows the differences between AI Red Teams and Cyber Red Teams by graphing the coverage of the literature reviewed by \citep{sinha2025what} across common stages of a red team engagement. Cyber Red Teams report better coverage of red-teaming stages, but none of the papers considered by the review noted a Cyber Red Team exploiting an AI component. In contrast, AI Red Teams do not report pre-engagement, scanning, vulnerability analysis, or cyber exploitation stages, but all papers considered noted an AI Red Team exploiting an AI component. The complementary approaches of AI and Cyber Red Teams suggest opportunities to better leverage each community's red-teaming strengths.

\textbf{We take the position that AI systems can be more effectively red-teamed if AI red-teaming is recognized as a domain-specific evolution of cyber red-teaming.} 
Figure~\ref{fig:cyber_ai_comparison} (right) summarizes how the two communities of practice may benefit from each other.
Cyber Red Teams augmented with additional AI expertise can utilize both the maturity of cyber red-teaming and the domain-specific knowledge of AI Red Teams to most effectively red-team AI systems, and AI Red Teams can more thoroughly test AI systems by also considering their non-AI components. 

In the remainder of this paper, we present an argument for our position and outline a path forward for merging AI and cyber red-teaming. Section~\ref{sec:ai-is-cyber} details our argument that AI red-teaming should be considered a domain-specific evolution of cyber red-teaming. Section~\ref{sec:evolving-cyber} describes the ways in which Cyber Red Teams should adapt to the changing AI threat landscape by leveraging AI red-teaming's expertise. Section~\ref{sec:lessons-for-ai} compares the current shortcomings of AI red-teaming with the corresponding strengths of Cyber Red Teams and provides recommendations on how to adopt existing cyber red-teaming best-practices.

\section{AI Red-Teaming is an Evolution of Cyber Red-Teaming}\label{sec:ai-is-cyber}

Our argument that AI red-teaming should be jointly conducted is motivated by current limitations of both fields in addressing system-level vulnerabilities in AI systems. Most AI systems, such as chatbots, self-driving cars, and AI agents, consist of one or more underlying AI models integrated into many layers of traditional software. Attackers do not delineate between arbitrary distinctions of AI vs. non-AI -- they exploit the weakest link in a system \citep{apruzzese2023real}. Red-teaming only AI components or only traditional software components in these systems fails to properly emulate adversaries -- the defining feature of red-teaming \citep{cnss2015glossary}. The best path towards achieving effective red-teaming of AI systems is by viewing AI red-teaming as a domain-specific evolution of cyber red-teaming and utilizing Cyber Red Teams augmented with additional AI expertise or AI Red Teams augmented with additional cybersecurity expertise. Cyber red-teaming has a mature ecosystem built over decades of experience in red-teaming critical systems which AI red-teaming can leverage, and the adoption of AI red-teaming will accelerate cyber red-teaming's evolution towards handling AI systems.

\paragraph{AI red-teaming needs the cyber red-teaming ecosystem.} Cyber red-teaming has accumulated decades of expertise in red-teaming critical systems into established red-teaming methodologies and best practices \citep{strom2018mitre, owasp_top10_2021, ISECOM_2012, herzog2003open, owasp2020guide, pci2024payment}; a strong community with standards, ethics, and procedures around safe and responsible red-teaming \citep{nist2024csf, householder2017cert, schaffer2023recommendations}; a diverse and mature ecosystem of tools \citep{tigner2021analysis, holik2014effective, sinha2025what, russo2016benefits}; and established institutions for informing and coordinating a diverse group of stakeholders \citep{householder2017cert, schaffer2023recommendations, cisa2024disclosure}. In contrast, AI red-teaming suffers from a lack of formalized red-teaming procedures \citep{longpre2024position}, proper adversary emulation \citep{sinha2025what}, responsible disclosure \citep{longpre2024position}, and mature red-team tooling \citep{sinha2025what}. Cyber red-teaming has established methodologies for handling each of these issues: rules of engagement (RoEs) \citep{ISECOM_2012}, structured threat modeling \citep{liu2012software}, CVD \citep{householder2017cert}, and tool libraries~\citep{tigner2021analysis, holik2014effective}, respectively. Where possible, it will be easier, faster, and cheaper to expand the existing ecosystem to encompass AI, rather than starting from scratch. 

\paragraph{AI red-teaming can be readily incorporated into the cyber ecosystem.} The line between an AI system and a software system is blurring as AI systems are growing to involve many layers of software that interact with an underlying AI model (e.g. \citep{badue2021self}). From the security side, a vulnerability in a complex AI system is unlikely to have a single point of failure in an underlying AI model, but rather stem from complex dependencies with software components \citep{householder2024lessons}. To maintain relevance, AI red-teaming will have to become increasingly system-focused, and therefore cybersecurity-focused. Of course, AI systems are notably distinct from traditional software in many ways, but these differences are not a deterrent to integration, as we discuss below. If AI systems are software systems, they are part of cyber red-teaming and there is no need to rebuild cyber institutions for the sake of AI security.

\paragraph{Cyber red-teaming will best adapt to AI systems by adopting AI red-teaming.} Regardless of where AI red-teaming sits in relation to cyber red-teaming, Cyber Red Teams will have to adapt to the consequences of the proliferation of AI. AI is reshaping the threat landscape, posing new risks and failure modes, and giving attackers and defenders new tools and assets with which to compromise and secure. However, cyber red-teaming is a broad framework that encompasses many unique technologies with unique behaviors and considerations. For previous major technological shifts, such as the widespread adoption of the Internet \citep{orman2003morris}, cloud services \citep{jansen2011guidelines}, and IoT devices~\citep{malhotra2021internet}, cyber red-teaming successfully both imposed structure and evolved to accept the new domain. The AI red-teaming community has the native AI expertise that cyber red-teaming requires, and incorporating AI into cyber will accelerate this evolution. The melding of AI and cyber red-teaming will not be immediate; it will take work and collaboration from both sides. However, the nature of the technology demands a comprehensive approach to security, covering both AI and traditional software components. The adoption of AI red-teaming by cyber red-teaming will best position the two fields to collaborate effectively.

\subsection{Alternative View}

An alternative view to our position is that software systems and AI systems are different in kind and therefore the red-teaming of each should focus on the core set of problems unique to each. Proponents of this view advocate for creating new, distinct CVD centers for coordinating the disclosure of AI vulnerabilities \citep{longpre2025house} and separate bug-bounty programs for AI vulnerabilities with different rules and expectations for AI red-teams \citep{bucknall2023structured}. Furthermore, proponents of this view argue that AI is sufficiently different from software such that boxing AI red-teaming into the realm of cybersecurity will stifle effectiveness and progress by imposing the wrong structure during a period of rapid change \citep{peterson2025insights}.

We believe that this is the wrong approach. Even if software systems and AI systems are different in kind, the vulnerabilities to each are not. For example, adversaries do not delineate between an `AI problem' or a `software problem', they follow the path of least resistance \citep{apruzzese2023real}. Furthermore, adversary emulation requires system-level thinking, encompassing software, AI, and the interaction between the two, and thus indicates a need for a combined approach to red-teaming. For example, systems whose components are individually safe and secure can still be vulnerable due to possible non-linear interactions of their components \citep{FRIEDBERG2017183, Avizienis2004}. Similarly, a secure system can be constructed from insecure components, provided that proper controls are in place \citep{Avizienis2004}.

\section{Evolving Cyber Red-Teaming in Response to AI}\label{sec:evolving-cyber}

AI is bringing about a new era of software that cyber red-teaming must address, and augmenting Cyber Red Teams with the expertise of AI red-teaming will best position Cyber Red Teams to evolve to the new threat landscape. AI is fundamentally reshaping the cybersecurity landscape by not only changing systems that must be secured, but also transforming the types of tools both red-teamers and adversaries have at their disposal. Not only do AI systems pose an array of new risks, but the risks also manifest through a range of new failure modes, demanding new threat models, red-teaming techniques, and mitigation prioritization. Though cyber red-teaming is well-poised to absorb these changes (just as it has with previous software revolutions), managing this change will still require considerable research, investment, and coordination from the AI and cyber red-teaming communities. Cyber red-teaming can best facilitate this transition by adopting the AI red-teaming community's expertise in managing the unique considerations of AI systems.

\subsection{AI Poses New Risks Beyond Those of Traditional Software}\label{sec:cyber-risks}

AI systems pose new risks that require new considerations to understand, emulate, and respond to threats. Cyber red-teaming has developed a sophisticated ecosystem for modeling cyber threat actors, using comprehensive shared frameworks \citep{strom2018mitre}, system telemetry \citep{xiang2021cisa}, and community threat intelligence sharing \citep{johnson2016guide}, but this ecosystem is focused on traditional cybersecurity risks. 

\begin{figure}[htb]
\centering
\includegraphics[width=0.95\textwidth]{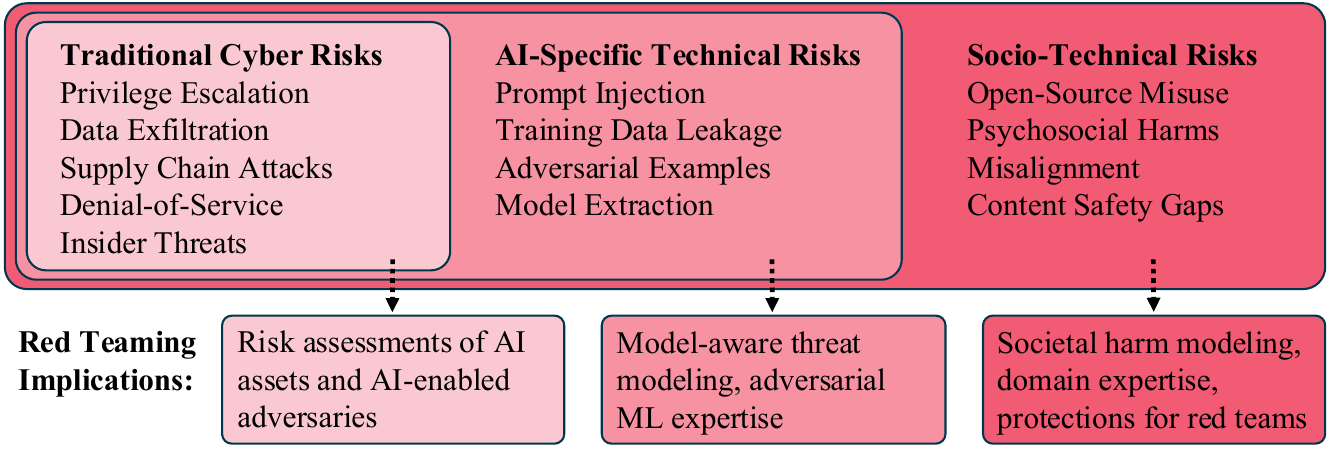}
\caption{A layered taxonomy of AI risks and their implications for red-teaming. As risks expand outward from traditional cybersecurity issues to AI-specific and socio-technical concerns, red-teaming methods must evolve accordingly.}
\label{fig:ai_risk_taxonomy}
\end{figure}

In Figure~\ref{fig:ai_risk_taxonomy} we categorize some of the risks of AI systems, including traditional cybersecurity risks of non-AI components, AI-specific technical risks, and socio-technical risks, and give a few implications these risks have on AI red-teaming. The AI community has been strongly invested in understanding these risks, and Cyber Red Teams should leverage their insights. 

\paragraph{AI exacerbates traditional cybersecurity risks.} The traditional cybersecurity risk taxonomy of confidentiality, integrity, and availability (CIA) \citep{cawthra2020data} remains applicable to AI systems, but with different information, data, and systems at risk. These risks include an increased risk of theft of intellectual property, such as training datasets, training methods, model architectures, and model weights; theft of sensitive data, such as personal medical information \citep{kretzschmar2019can, dave2023chatgpt}; and the disruption of processes such as training or inference pipelines. Furthermore, AI-enabled cyberattacks lower the barrier to entry for executing cyberattacks against software systems. Real-world cyber threat actors are turning to AI systems to help with or automate parts of the attack chain \citep{gtig2025misuse, rodriguez2025framework}. The continued improvement of AI capabilities could profoundly change the cybersecurity threat landscape \citep{gtig2025misuse}.

\paragraph{AI models have unique technical risks that Cyber Red Teams must address.} There are an array of risks specific to AI models that require domain-specific knowledge to address. These include adversarial examples \citep{szegedy2013intriguing}, membership inference \citep{shokri2017membership}, model extraction \citep{tramer2016stealing}, training data extraction~\citep{carlini2021extracting}, prompt injections \citep{greshake2023not}, reward hacking \citep{skalse2022defining}, data poisoning \citep{biggio2012poisoning}, and more. The potential harms from these risks can map to both traditional cybersecurity risks (e.g., membership inference attacks risk loss of confidentiality), and to AI-specific socio-technical risks.

\paragraph{AI systems pose socio-technical risks beyond the scope of typical cybersecurity risks.} Traditional cyber red-teaming is primarily concerned with harms to organizations \citep{ISECOM_2012}, such as economic or reputational harms. AI systems, however, pose a variety of risks deeply intertwined with societal or human factors. Recent advances in generative AI have raised concerns over psychosocial harms such as content safety and mental health, which have led to AI red-teaming efforts to avoid such behaviors~\citep{weidinger2023sociotechnical, bullwinkel2025lessons}. AI systems even present risks for the red-teamers themselves, in ways similar to those for content safety moderators \citep{zhang2024human}. AI models can be misused to cause harm to other systems~\citep{weidinger2023sociotechnical}, for example by automating cyber attacks \citep{rodriguez2025framework}. Open-source models complicate the handling of misuse risks, as developers must defend against adversaries with full white-box access, and vulnerabilities cannot be mitigated after model release \citep{seger2023open, eiras2024near, bommasani2024considerations}. Advances in agentic AI capabilities have led to increasing concerns about misalignment of AI systems, or systems that operate autonomously in ways that do not reflect the intentions of developers \citep{shah2025approach}. This differs substantially from existing cybersecurity threat models, as the software itself is presented as the adversary, although parallels can be drawn to insider threat modeling \citep{greenblatt2023ai}. Misaligned systems currently appear to pose very little risk \cite{krakovna2020specification, bondarenko2025demonstrating} or only demonstrate misalignment in artificial scenarios \citep{greenblatt2024alignment, meinke2024frontier}, but this may be an increasingly relevant concern as AI agents become more capable and interact more with the real world.

\paragraph{Cyber Red Teams should consider the additional risks of AI systems.} Accurate enumeration of high-risk vulnerabilities will require an up-to-date understanding of the expanded attack surface and how threat activity and offense-defense balances have shifted across that attack surface. We suggest that Cyber Red Teams take the following additional steps when red-teaming AI systems:
\begin{itemize}
    \item Quantify the business or mission value of AI system assets and their likelihood of exploitation in the pre-engagement phase.
    \item If the goals of the engagement include socio-technical risks, consult with relevant domain experts to accurately assess risks, such as psychologists or sociologists for psychosocial harms, and chemists or virologists for AI-assisted weapon development. 
    \item If psychosocial harms are considered within the engagement, take steps to avoid exposing red-teamers to content safety risks and provide recovery services.
\end{itemize}

\subsection{AI Has New Failure Modes, Requiring New Red-Teaming Tactics to Trigger Them}

The risks posed by AI systems can manifest through failure modes beyond those of traditional software. New red-teaming techniques and methodologies are therefore needed to effectively expose and mitigate vulnerabilities. We identify the domain, emergent behavior, and opacity of AI systems as the core properties that distinguish AI systems from traditional software. Augmenting Cyber Red Teams with AI-specific red-teaming expertise will promote the development of tools and techniques that better address risks stemming from these properties of AI systems. 

\paragraph{Domain.} AI has seen the most development in domains where traditional software struggles to provide value, such as classifying \citep{deng2009imagenet} or detecting objects \citep{lin2014microsoft}, translating \citep{barrault2019findings} or generating text \citep{zhu2015aligning}, and autonomously operating games \citep{silver2016mastering} or machines \citep{pomerleau1988alvinn}. Tools will need to be developed or improved to handle the unique requirements of these domains and interface with existing tooling. For example, the introduction of autonomous vehicles has required cybersecurity to adapt practices to that domain~\citep{cisa2021autonomous}, such as the development of virtual simulators to assess the collision risks posed by cyberattacks on vehicle communication networks \citep{izlar2023simulating}.

\paragraph{Emergence.} While there has been much debate around the nuances of `emergent' abilities in AI models \citep{wei2022emergent, schaeffer2023emergent}, it is clear that AI models routinely succeed and fail in ways that humans fail to predict. Failures that only emerge in rare or critical situations pose challenges for red-teaming as it is often unclear how to go about finding them. For instance, failures may come from specific pieces of data whose impact only manifests down the pipeline \citep{goldblum2022dataset, price2024future}. In comparison, common non-AI vulnerabilities tend to be the result of specific isolated mistakes, such as not sanitizing input data or misconfiguring access controls \citep{owasp_top10_2021}. 

\paragraph{Opacity.} When AI failures occur, it is often difficult to understand why that particular failure occurred. AI models are popularly described as `black-boxes' \citep{rudin2019we}, where our understanding only comes from input-output behavior, rather than the internal mechanisms used to produce the outputs. Although progress has been made on interpreting the behavior of model internals, many open problems remain \citep{sharkey2025open}. Traditional software generally does not have this issue, as the source code is designed to be human-understandable. However, there are cases where software functions as a black-box, such as closed-source APIs, binaries, or poorly documented source-code.

\paragraph{Cyber Red Teams should incorporate new red-teaming techniques to fully assess the AI components of systems under test.} As AI systems blur the boundary between software behavior and learned policy, additional techniques are useful for Cyber Red Teams to develop. Areas for consideration include: 
\begin{itemize}
    \item Adapt and develop red-teaming tools to increase red-team effectiveness in AI-specific domains such as machine vision, natural language, and autotomy.
    \item Design engagements that can capture behavior under uncertainty or nondeterminism (e.g., randomized testing, varying environmental conditions, adversarial probing).    
    \item Develop long-tail testing frameworks for rare issues and emergent behavior.
\end{itemize}

\subsection{AI Re-prioritizes Disclosure and Mitigation Strategies}

Persistent, unpatchable AI vulnerabilities pose challenges which require re-prioritization of careful disclosure protocols and comprehensive mitigation strategies during the cyber red-teaming lifecycle.

\paragraph{AI has unpatchable vulnerabilities.} Many of the most well-known vulnerabilities in AI systems lack known fixes \citep{hendrycks2021unsolved}. Adversarial examples on image classifiers \citep{szegedy2013intriguing}, for example, have been studied for over a decade yet little progress on robustness has been made \citep{croce2020robustbench}. In contrast, cybersecurity vulnerabilities are often addressed with small code changes.

\paragraph{Cyber also contends with unpatchable vulnerabilities.} While most cybersecurity vulnerabilities can be addressed with feasible solutions, the field does have notable examples of long-lived, unpatchable vulnerabilities. For example, Spectre is a class of vulnerabilities in modern processors that is considered to be unpatchable in general \citep{kocher2020spectre, mcilroy2019spectre}, and the Internet's Border Gateway Protocol (BGP) is an inherently insecure protocol that is too entrenched to replace \citep{rekhter2006border, ballani2007study, haag2019protecting}. In these kinds of cases, the industry's approach is to patch specific known exploits where possible, implement safeguards that increase attack costs, and maintain an overall defense-in-depth approach~\citep{varda2020mitigating, kocher2020spectre, lipp2020meltdown, zhang2007practical}.

\paragraph{AI vulnerabilities require greater caution.} 
To secure AI systems, Cyber Red Teams and system owners will need to develop methodologies to assess the systemic impact of vulnerabilities that are unpatchable or time-consuming to mitigate, test secure-by-design assumptions earlier in the development lifecycle~\citep{cisa2023secure}, and document mitigation pathways beyond conventional patching \citep{mell2016measuring}. Unlike the Internet's dependence on BGP \citep{haag2019protecting}, AI systems are not yet locked into insecure legacy processes, offering a rare opportunity to standardize around secure architectures before insecure practices take hold. Realizing this opportunity will require coordination across the AI and cybersecurity communities and may necessitate regulatory or incentive-driven interventions to ensure adoption of robust practices.

\paragraph{Cyber Red Teams should change reporting to accommodate unpatchable AI vulnerabilities.}
\begin{itemize}
    \item Promote secure design practices that do not rely on perfect security of AI models.
    \item Emphasize defense-in-depth to layer security around insecure AI models and raise attack costs.
    \item Treat unpatchable AI vulnerabilities as \textit{classes} of vulnerabilities, and specific implementations of these attack classes as distinct vulnerabilities, similar to CWEs vs. CVEs.
    \item Prepare CVD protocols to manage unpatchable vulnerabilities, especially in open-source models. These may require extended disclosure timelines and greater coordination among more stakeholders. 
\end{itemize}

\section{Evolving AI Red-Teaming in Response to Cyber Red Team Best Practices}\label{sec:lessons-for-ai}

The relatively recent development of AI red-teaming has left little time to develop and formalize best-practices, tools, or methodologies. We argue that many of the current limitations of AI red-teaming are already addressed by current cyber red-teaming practices. In this section, we suggest ways that an existing AI Red Team may be able to accomplish more of their goals by incorporating Cyber Red Team best practices. 

Specifically, we suggest three areas where AI Red Teams can draw from their cybersecurity colleagues. First, AI red-teams lack detailed threat modeling and adversary emulation \citep{sinha2025what}, leading to outcomes that potentially do not address real threats to systems \citep{mazeika2024harmbench}. Second, AI red-teaming fails to establish shared expectations for red-teaming and disclosure among various stakeholders \citep{longpre2024position, longpre2025house, sinha2025what}, which reduces mutual accountability. Finally, AI Red Teams lack mature tooling as the open-source tools that exist are largely repositories of research code. 

\subsection{Adversary-based Threat Modeling Can Help Discover High-risk Vulnerabilities}\label{sec:ai-threat-models}

AI systems must be red-teamed as \emph{systems}, not as AI models. This requires consideration of the entire risk surface, including, and especially, traditional cyber threats \citep{apruzzese2023real}. AI red-teaming has been criticized for overly focusing on the underlying AI model and failing to follow realistic threat models or emulate real-world adversaries \citep{apruzzese2023real, sinha2025what}. As cyber red-teaming already places major emphasis on adversary emulation and threat modeling \citep{ISECOM_2012, xiong2019threat, sinha2025what}, viewing AI red-teaming as an evolution of cyber red-teaming naturally motivates more comprehensive evaluation of AI systems. 

\paragraph{Cyber red-teaming uses structured threat modeling to target high-risk vulnerabilities.} The cybersecurity risk surface is extremely broad, and red-teaming, by nature, is resource intensive \citep{teichmann2023overview}. It is therefore crucial that Cyber Red Teams prioritize high-risk vulnerabilities to maximize impact. Cyber red-teaming uses a structured approach to threat modeling, explicitly modeling the goals, resources, and Tactics, Techniques, and Procedures (TTPs) \citep{johnson2016guide} of specific threat actors or archetypes, in order to emulate likely attack behaviors \citep{ISECOM_2012, xiong2019threat, sinha2025what}, thereby identifying vulnerabilities that genuine threat actors are most likely to exploit. This does not preclude the use of speculative threat models to discover emerging attack techniques, especially for researchers, but simply demands that threat models be comprehensive and grounded in reality, and that engagements emulate the threat model accurately \citep{saeed2023systematic}.

\paragraph{AI red-teaming places little emphasis on practical threat modeling.} AI Red Teams often fail to engage rigorously in practical threat modeling (c.f.  Figure~\ref{fig:cyber_ai_comparison}), with common failures including focusing solely on the AI model, ignoring easier paths to the same end, or speculating on threat models without grounding them in real-world threat intelligence. Recent AI security research on jailbreaks (forcing generative AI models to generate content against their safety policies) has been criticized for failing to consider alternative ways the content could be produced or found (e.g., via a web search \citep{mazeika2024harmbench}), leaving threat models implicit, and prioritizing marginal gains in attack success rate rather than downstream impact \citep{sinha2025what, rando2025do}. In generative AI more broadly, there has been some criticism of the focus on future AI risks, rather than current concerns \citep{editorials2023stop, bender2023ai}. 

\paragraph{AI Red Teams should adopt structured threat modeling for threat prioritization.} The breadth of risks posed by AI systems and the limited resources available to address them suggest a prioritization through structured threat modeling. We recommend to: 
\begin{itemize}
    \item Define an explicit threat model with as much specificity as possible and adhere to it closely.
    \item Develop threat actor profiles based on threat intelligence and system telemetry, and update these threat actor profiles frequently to reflect new intelligence, telemetry, and research.
    \item Target the highest risk vulnerabilities by weighing the quantified risk and likelihood of exploitation.
\end{itemize}

\subsection{Mutual Accountability Enables Impactful Red Team Outcomes}\label{sec:ai-accountability}

AI red-teaming has been criticized for a lack of accountability among stakeholders~\citep{longpre2024position,sinha2025what}. We recommend AI Red teams leverage cyber red-teaming's existing culture of mutual responsibility to promote more effective red-teaming. Red-teaming requires investment, coordination, and trust from many disparate parties who must hold each other responsible for the security of a system. In particular, it relies on close coordination between the red-team and the host organization to define the goals, threat model, RoEs, and reporting requirements \citep{sinha2025what}. Critics highlight a lack of trust between hosts and red-teams \citep{longpre2024position}, as well as a lack of clear reporting and disclosure procedures \citep{longpre2024position, sinha2025what}. These failures demonstrate a lack of mutual accountability which, if left unresolved, may stifle collaboration and good-faith efforts to red-team AI systems. 

\paragraph{Rules of engagement manage risk and influence outcomes.} Cyber Red Teams take care to collaborate with the host organization to define the RoEs prior to a red-teaming engagement, which define the scope, threat model, rules, and legal considerations of the engagement \citep{ISECOM_2012, sinha2025what}. RoEs determine the way red-teams operate by prioritizing certain risks, methodologies, team dynamics, expertise, and tools, all of which impact the results of the engagement~\citep{feffer2024red}. They also balance incentives and risks for each party, as well as other stakeholders, by setting safeguards and establishing red lines for red-teamers to prevent incidental losses \citep{herzog2003open, gupta2024conceptual}; defining expectations for disclosure, mitigation, and publication timelines \citep{householder2017cert, longpre2025house}; and providing legal protections for good-faith red-teaming \citep{demarco2018approach, longpre2024position}.

\paragraph{Reporting will alert affected parties and inform mitigations.} Cyber red-teaming has a strong reporting culture \citep{walshe2022coordinated}, reflecting the importance of communicating vulnerabilities to motivate mitigations~\citep{shah2015overview}. Reporting is viewed as a joint effort between Cyber Red Teams and host organizations, where the red-team is responsible for demonstrating the vulnerability and fairly advocating for the risk it poses, and the host is responsible for collaborating with the red-team to develop and implement mitigations~\citep{smith2020case, scarfone2008technical, spring2021prioritizing}. Depending on the situation, one of the parties will also be responsible for following CVD protocols \citep{householder2017cert, householder2024lessons}. Outside of specific engagements, organizations are expected to share threat intelligence to inform threat models \citep{johnson2016guide}.

\paragraph{AI Red Teams often do not have explicit accountability mechanisms in place.} Despite the emphasis on coordination between host and red-team in cyber red-teaming, AI red-teaming has not yet exhibited the same commitment. The usage policies of leading AI model providers often do not provide protection for good-faith security research \citep{longpre2024position}. Security researchers often fail to engage in standard responsible disclosure practices or fail to follow practical threat models \citep{sinha2025what}. Although prior work has advocated for CVD in AI red-teaming~\citep{longpre2025house}, we emphasize that the success of CVD processes depends on mutual accountability between parties across the security ecosystem to both disclose and be receptive of vulnerabilities. For example, a vulnerability for extracting training data discovered by \citet{nasr2025scalable} was disclosed to, and patched by, OpenAI, yet Google later released models with the same vulnerability \citep{nasr2025scalable}, a situation which could potentially have been avoided with better coordination. Organizations must be willing to both share vulnerabilities and implement mitigations for CVD to be effective, and avoid similar situations in the future. 

\paragraph{AI Red Teams should adopt accountability mechanisms to build trust with hosts.}
In Table~\ref{tab:mutual-accountability} we compare the responsibilities of the host organization and red-team, which highlights the joint nature of the red-teaming process and the mutual accountability of both.

\begin{table}[htb]
\renewcommand{\arraystretch}{1.2}
\caption{Mutual responsibilities of host organizations and red-teams.}
\label{tab:mutual-accountability}
\small
\centering
\begin{tabular}{|p{0.465\linewidth}|p{0.465\linewidth}|}
\hline
    \textbf{Host Responsibilities} & \textbf{Red-Team Responsibilities} \\ \hline

    Define in the RoEs the scope and desired outcomes, threat profiles and prioritization, and desired artifacts to inform mitigations (reports, demos, etc.). & Inform the RoEs with prior experience. Prioritize the host's goals and follow the RoEs and chosen threat models. \\ \hline
    
    Grant reasonable legal protections to good-faith red-teaming via bug bounties and research exemptions. & Red-team in good faith and stay within the established bounds. \\ \hline

    Establish responsible disclosure policies and permit publishing vulnerabilities following a grace period. & Communicate vulnerabilities in accordance with responsible disclosure policies. \\ \hline

    Establish which parties are responsible for CVD. If responsible, follow established CVD protocols. Update the red-team on the status of disclosure. & If responsible, follow established CVD protocols. If not, consider following up on the status of disclosure. \\ \hline
\end{tabular}
\renewcommand{\arraystretch}{1}
\end{table}

\subsection{Maturing AI Red-Teaming Tools Will Lower the Barrier to Effective Engagements}\label{sec:ai-tools}

AI red-teaming could greatly benefit from a mature, comprehensive set of red-teaming tools. Cyber red-teaming employs a comprehensive suite of mature tools, many open-source, which, despite dual-use concerns, consistently promote a defender's advantage by better allowing vulnerabilities to be patched prior to deployment. Popular programs like Kali Linux \citep{tigner2021analysis} and Metasploit \citep{holik2014effective} centralize a large number of tools, allowing many different tools to be used in conjunction. In contrast, however, the maturity of open-source AI red-teaming tools, and their integration with existing cyber red-teaming tools, has lagged behind the development of AI systems \citep{sinha2025what}.

\paragraph{Cyber red-teaming's mature tool ecosystem promotes security.} Cyber red teams employ a diverse selection of well-known, well-maintained, and widely used tools, which automate or reduce the complexity of common tasks \citep{sinha2025what}. The breadth of options in tool suites like Kali Linux and Metasploit reflect cyber red-teaming's comprehensive system-level approach to red-teaming (Figure~\ref{fig:cyber_ai_comparison}) \citep{holik2014effective, tigner2021analysis}. Many tools are open-source \citep{sinha2025what}, and although there are legitimate concerns about the malicious use of these tools \citep{agcaoili2021locked, gallagher2025sophos}, they ultimately provide a defender's advantage by making vulnerability discovery repeatable and efficient \citep{sinha2025what}. Cyber red-teaming emphasizes the importance of using tools to mitigate `low-hanging fruit' before considering more complex vulnerabilities, since it reflects adversaries' behavior \citep{owasp_top10_2021, martin20112011}.

\paragraph{AI red-teaming has a lack of mature tools.} The AI red-teaming community, on the other hand, does not have a comparable tool ecosystem \citep{sinha2025what}. Red-teaming tools exist largely as research code and are often neither regularly maintained nor designed to be scalable for production-level red-teaming. The tools are also highly model-centric and ignore much of the software surrounding the underlying model in production systems, reflecting the lack of threat modeling discussed in Section~\ref{sec:ai-threat-models}. While the lack of mature AI red-teaming tools is perhaps understandable given the newness of the technology, it does not reflect the growth of production-level AI systems and is an indicator of security lagging behind capabilities. There are corporate solutions being developed (see, e.g., \citep{lakera2025red, thigpen2025red}), but their level of adoption is unclear and do not replace the need for a robust open-source ecosystem.

\paragraph{AI Red Teams should mature the red-teaming tool ecosystem by sharing the tooling they develop.} Specific recommendations include:
\begin{itemize}
    \item Develop open-source tooling by contributing to existing projects or publicly releasing new tools.
    \item Build interoperability between AI-specific tools and existing cyber red-teaming tools in order to share out with red teams of various levels of AI expertise. 
    \item Encode red-teaming best-practices such as logging, reproducibility, and safety controls, within the tooling developed to support engagements. 
\end{itemize}

\section{Conclusion}\label{sec:conclusion}

AI red-teaming should be understood as a domain-specific evolution of cyber red-teaming: grounded in the same foundational practices but expanded to address the distinct challenges introduced by AI systems. This framing strengthens both fields, bringing structure, accountability, protections, and tool-driven rigor to AI red-teaming while pushing cybersecurity to evolve its practices in response to AI-specific risks. A shared framework avoids fragmenting the security landscape, supports consistent evaluation standards, scales tools and workflows across domains, and enables more effective coordination across research, industry, and policy. Most importantly, it ensures red-teaming resources are focused on the most consequential vulnerabilities spanning both AI and traditional software systems.

We call for closer collaboration between the AI and cybersecurity communities to develop integrated red-teaming programs that combine expertise in AI systems, security engineering, software infrastructure, and threat modeling. Cyber Red Teams must begin treating AI systems as core components of the modern attack surface. Future work should aim to formalize hybrid threat modeling, build frameworks that encompass both traditional and AI-specific vulnerabilities, situate hybrid red-teaming within AI governance frameworks, and develop lifecycle-aware evaluation processes that keep pace with AI's rapid evolution. A unified red-teaming discipline is essential for securing the next generation of software systems.

\section{Acknowledgments}

Copyright 2025 Carnegie Mellon University.

This material is based upon work funded and supported by the Department of Defense under Contract No. FA8702-15-D-0002 with Carnegie Mellon University for the operation of the Software Engineering Institute, a federally funded research and development center.  

The view, opinions, and/or findings contained in this material are those of the author(s) and should not be construed as an official Government position, policy, or decision, unless designated by other documentation.

NO WARRANTY. THIS CARNEGIE MELLON UNIVERSITY AND SOFTWARE ENGINEERING INSTITUTE MATERIAL IS FURNISHED ON AN "AS-IS" BASIS. CARNEGIE MELLON UNIVERSITY MAKES NO WARRANTIES OF ANY KIND, EITHER EXPRESSED OR IMPLIED, AS TO ANY MATTER INCLUDING, BUT NOT LIMITED TO, WARRANTY OF FITNESS FOR PURPOSE OR MERCHANTABILITY, EXCLUSIVITY, OR RESULTS OBTAINED FROM USE OF THE MATERIAL. CARNEGIE MELLON UNIVERSITY DOES NOT MAKE ANY WARRANTY OF ANY KIND WITH RESPECT TO FREEDOM FROM PATENT, TRADEMARK, OR COPYRIGHT INFRINGEMENT.

[DISTRIBUTION STATEMENT A] This material has been approved for public release and unlimited distribution.  Please see Copyright notice for non-US Government use and distribution.

This work is licensed under a Creative Commons Attribution-NonCommercial 4.0 International License.  Requests for permission for non-licensed uses should be directed to the Software Engineering Institute at permission@sei.cmu.edu.

DM25-1040

\bibliography{neurips_2025}
\bibliographystyle{plainnat}

\end{document}